\documentclass[a4paper]{jpconf}
\usepackage{graphicx}
\usepackage[english]{babel}  
\usepackage[usenames]{color} 

\begin{document}

\noindent \textcolor{red}{
	International Conference ``Intelligent Systems in Science and Technology''.
	The 5th All-Russian Scientific and Practical Conference ``Artificial Intelligence in Solving Actual Social and Economic Problems of the 21st Century'', Perm State National Research University (Perm, Russia), October 12-18, 2020. }

\noindent \textcolor{red}{
	The text of the talk will be published in the {\bf Journal of Physics: Conference Series}. }

\title{Application of computer simulation results and machine learning in analysis of microwave radiothermometry data}

\author{Maxim Polyakov$^{1}$, Illarion Popov$^{1}$, Alexander Losev$^{1}$, Alexander Khoperskov$^{1}$}

\address{Volgograd State University, 100, Prospect Universitetsky, Volgograd, Russia}

\ead{\{m.v.polyakov, popov.larion, alexander.losev, khoperskov\}@volsu.ru}

\begin{abstract}
This work was done with the aim of developing the fundamental breast cancer early differential diagnosis foundations based on modeling the space-time temperature distribution using the microwave radiothermometry method and obtained data intelligent analysis. The article deals with the machine learning application in the microwave radiothermometry data analysis. The problems associated with the construction mammary glands temperature fields computer models for patients with various diagnostics classes, are also discussed. With the help of a computer experiment, based on the machine learning algorithms set (logistic regression, naive Bayesian classifier, support vector machine, decision tree, gradient boosting, K-nearest neighbors, etc.) usage, the mammary glands temperature fields computer models set adequacy.
\end{abstract}

\section{Introduction}
\noindent Currently, mathematical modeling and machine learning methods are actively used in developing new medical diagnostics systems. In particular, this approach provides the opportunity to create effective functional diagnostics methods based on the measurement, description and interpretation of the human body physical field parameters. However, the medicine instrumental base development has led to the fact that problems in the diagnostic solution formulation in most cases arise not due to a lack of information, but due to insufficient methods efficiency for its processing. The problems solution provides the systems creation for the medical data interpretation and analysis \cite{Yasnitsky}. Such systems, using machine learning methods and algorithms, should help specialists in making diagnoses and predicting the diseases development \cite{Levshinskii1}.

At the same time, one of the most difficult and urgent medicine problems is the breast diseases early diagnosis problem. Breast cancer is the most common cancer among women. The solution to the increasing survival through early disease detection seems to be quite obvious. Promising methods for increasing breast cancer early diagnosis efficiency is the microwave radiothermometry method \cite{Vesnin}.

This study was carried out within the direction framework, whose main goal is to develop the breast diseases early differential diagnosis fundamental foundations based on modeling the space-time temperature distribution using the microwave radiothermometry method and obtained data intelligent analysis.

It is known that human tissues, like any heated body, emit electromagnetic oscillations in a wide frequency range. In this case, the radiation spectrum and the tissue's own electromagnetic radiation intensity in the microwave range is determined by both the temperature distribution and the absorption and reradiation in physically inhomogeneous tissues. The situation is complicated by the fact that at the human body temperature, the maximum radiation falls on the near infrared range, the spectrum, which is formed on the biological tissue surface. And the information about the internal temperature distribution at a several centimeters depth is due to longer wavelength radiation, which is magnitude weaker orders. Thus, by measuring this radiation, we obtain extremely important information about the internal tissues and skin stat. This is the main microwave radiothermometry (RTM) method idea, based on the tissue's intrinsic radiation measurement in the microwave and infrared (IR) wavelength ranges.
Over the past two decades, this method has become widespread in various medicine fields \cite{Losev}. At the same time, it’s practical application analysis has identified a number of problems that require solutions. First of all, the need arose to develop adequate computer physical and mathematical models for studying the spatial and temporal temperature fields dynamics in the biological mammary gland tissues. In recent years, during theoretical research, a number of mathematical models have been created that describe the temperature distribution in human organs \cite{Losev, Polyakov1, Levshinskii2, Polyakov2}. Including in works \cite{Polyakov1, Levshinskii2, Polyakov2}, when modeling, the main macroscopic factors determining thermal dynamics were taken into account. Instead of traditionally used models with homogeneous parameters in a multilayer approximation (usually limited to four tissue types - skin, muscles, mammary gland, tumors), a close to realistic geometric tissues structure with heterogeneous characteristics was used in the simulation. The developed models also took into account filamentary connective tissues, breast lobules, nipple, excretory ducts, and adipose tissue. Note that a large number of input physical parameters seriously complicates the modeling process, but, at the same time, significantly increases the model adequacy. In real conditions, the parameters values spread lies within very wide limits. Also in the study, based on in-depth medical data analysis and using machine learning algorithms, an algorithm for validating a simulation model was proposed \cite{Levshinskii2}.

On the other hand, in recent years, the possibilities of using machine learning algorithms in the formulation and a diagnostic solution based on microwave radiothermometry data substantiation have been actively studied \cite{Levshinskii1, Losev, Levshinskii3, Losev2}. It was quickly established that an attempt to make a diagnosis using artificial intelligence methods based solely on temperature data does not provide the required sensitivity and specificity. Thermometric feature spaces various versions have been proposed \cite{Levshinskii1, Levshinskii3}. Their construction is based on a descriptive hypothesis set about the temperature fields behavior in patients of different diagnostic classes. In particular, for patients with pathology, there is an increased thermoasymmetry value between the same mammary glands points; increased temperature spread between individual points in the affected mammary gland; increased nipple temperature in the affected mammary gland compared to the average breast temperature, taking into account age-related changes in temperature; the skin and depth temperatures ratio and some others. During the study, the quantitative thermometric factors characteristics were discovered and described. A number of classification algorithms were built on the attribute spaces basis. In particular, the most efficient classification algorithms based on logistic regression showed sensitivity and specificity in the region of 0.65 - 0.7 \cite{Levshinskii3}. As expected, the artificial neural networks usage gives a fairly high efficiency. However, here, too, specific input signal preprocessing based on the corresponding feature spaces \cite{Levshinskii3} made it possible to significantly increase the accuracy. Namely, the combined modeling functions and the anamnesis results usage in the first layer, proposed in \cite{Losev2}, made it possible to increase the classifier efficiency by more than 10 percent compared to using only temperature values in the input layer \cite{Levshinskii1}. The sensitivity and specificity in the artificial neural network model proposed in \cite{Losev2} reached values of 0.8 - 0.85.

\section{Mathematical and numerical models}

We simulate temperatures on the surface and inside the breast. Our heat dynamics model is based on the Pennes equation \cite{Pennes} in a three-dimensional setting

\begin{equation}\label{eq-heat-dynamics}
 \varrho(\vec{r}) c_p(\vec{r})\, \frac{\partial T}{\partial t} = \vec{\nabla} \left( \lambda(\vec{r}) \vec{\nabla} T  \right) + \sum\limits_{j} Q_k(\vec{r})  \,,
\end{equation}

\noindent where $T(\vec{r})$  is the temperature at the point $\vec{r}=\left\{ x,y,z \right\}$, $\vec{\nabla}$ is differential operator nabla, heat source $Q_k > 0$ for $k=\{ met, can, ... \}$ and $Q_{rad}<0$, $T_{air}$ is temperature of environment, $S_0(\vec{r})$ is the boundary of biological tissue, $\vec{n}$ is unit normal vector. The equation specifies the boundary conditions between biological tissue and environment

\begin{equation}\label{eq-heat-dynamics-boundary}
\vec{n}(\vec{r})\cdot \vec{\nabla}T= \frac{h_{air}}{\lambda(\vec{r})} \cdot \left( T - T_{air} \right).
\end{equation}

\noindent Let us determine initial temperature distribution in order to reveal uniqueness of solution

\begin{equation}\label{initial-temp}
T|_{t=0}=T_0(\vec{r}).
\end{equation}

The finite element method is used to calculate spatial distribution of the thermodynamic temperature. We find solution to the problem in form of an expansion in system of basis functions $\{\gamma_j\}_{j=1}^N$

\begin{equation}\label{basis-fun}
T_h(\vec{r},t)=\sum\limits_{j=1}^N T_j(t)\gamma_j(\vec{r}).
\end{equation}

\noindent After transformations, we get a system of ordinary differential equations

\begin{equation}\label{ode}
\sum\limits_{j=1}^N\left(\frac{\partial T_j}{\partial t} \int\limits_{V_0}\varrho(\vec{r}) c_p(\vec{r})\gamma_i \gamma_j dV + T_j \int\limits_{V_0} \lambda(\vec{r}) \nabla\gamma_i\cdot \nabla\gamma_j dV \right) - \int\limits_{V_0} Q_k\gamma_i dV=0, i=1, ..., N,
\end{equation}

\noindent or in matrix form

\begin{equation}\label{matrix-form}
[G]\frac{\partial}{\partial t}\{T\}+[M]\{T\}=\{P\},
\end{equation}

\noindent where $G_{ij}=\int\limits_{V_0}\varrho(\vec{r}) c_p(\vec{r})\gamma_i \gamma_j dV$, $M_{ij}=\int\limits_{V_0} \lambda(\vec{r}) \nabla\gamma_i\cdot \nabla\gamma_j dV$ and $P_{i}=\int\limits_{V_0} Q_k\gamma_i dV$.

In finite element method terminology, [M] is the stiffness matrix and [G] is the damping matrix. At the first stage, the region $V_{0}$ is divided into finite elements $V_e$. After that, local basis functions are selected in each element. The implicit scheme for this problem is

\begin{equation}\label{n-scheme}
[G]\frac{\{\hat{T}\}-\{T\}}{\tau}+[M]\{\hat{T}\}=\{P\}.
\end{equation}

We use a diagonal damping matrix for this scheme. To obtain the values of the nodal temperatures at the next time layer, we solve a system of linear algebraic equations

\begin{equation}\label{slae}
[A]\{\hat{T}\}=\{b\},
\end{equation}

\noindent where

\begin{equation}\label{slae-r}
[A]=\frac{1}{\tau}[G^{diag}]+[M], \{b\}=\{P\}+\frac{1}{\tau}[G^{diag}]\{T\},
\end{equation}

\noindent where $[G^{diag}]$ is a diagonal damping matrix.

If a boundary condition of  first kind is specified on a part boundary of the region, then temperature at corresponding sides is determined by equation $\hat{T_{i}}=T_1(\vec{r},t)$. We used a mesh of tetrahedrons as finite elements.

\begin{figure}[h]
\begin{minipage}{38pc}
\includegraphics[width=38pc]{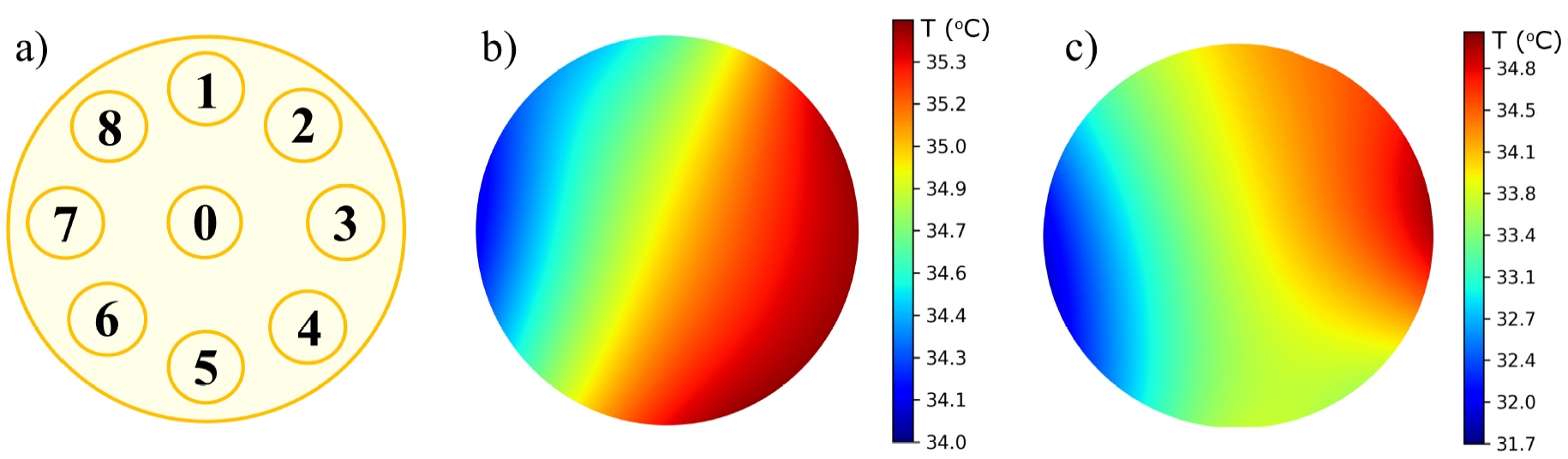}
\caption{\label{Temp} a) Scheme used in simulation of brightness temperatures for right breast. b) Internal temperature distribution for a model with a 1 cm radius tumor located at point "3". c) Surface temperature distribution for a model with a tumor with a radius of 1 cm located at point "3".}
\end{minipage}
\end{figure}

We define a complex circulatory network, in contrast to the classical biothermal equation, in which blood flows are evenly distributed throughout the entire volume of biological tissue. The blood vessels are heat sources in our model, and they have a temperature of 37 $^\circ$C. The radio-microwave thermometry method is based on the concept of brightness temperature \cite{Sedankin}. The brightness temperature is different from the thermodynamic temperature. Therefore, to build an adequate simulation model of the process of measuring the temperature of biological tissue, it is necessary to use equation

\begin{eqnarray}
 T_B^i =  \int\limits_{V_{0}} T(\vec{r}) \,\frac{P_d(\vec{r},f)}{\int_{V_0}P_d(\vec{r},f)\,dV}\,dV,
\end{eqnarray}

\noindent where $\displaystyle P_d = \frac{1}{2}\,\sigma(\vec{r},f)\cdot \left| \vec{E}(\vec{r},f) \right|^2$ is electromagnetic power density, $\vec{E}$ is electric field vector, $i\in [0,8]$ are points according to examination method for right breast (see fig. \ref{Temp}(a)).

\begin{figure}[h]
\includegraphics[width=22pc]{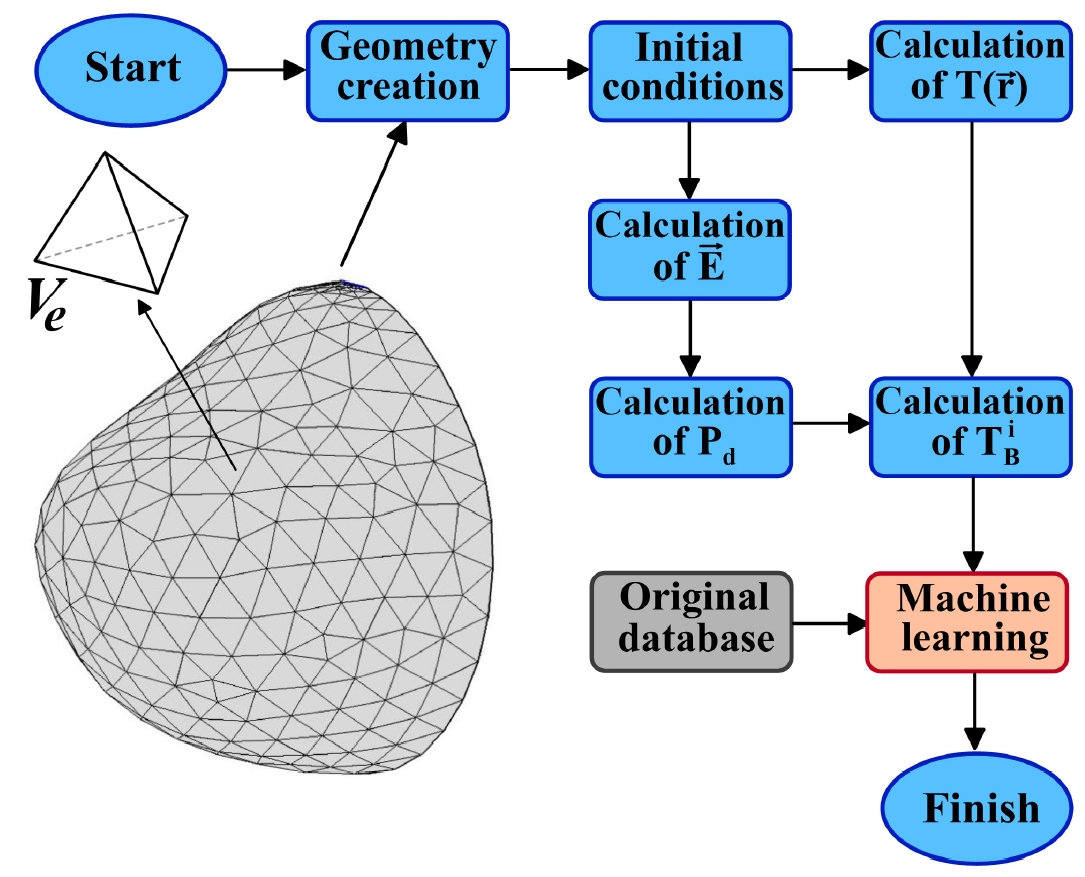}\hspace{2pc}%
\begin{minipage}[b]{14pc}\caption{\label{Algorithm}Algorithm for calculating brightness temperature in  mammary gland and post-processing. In the first step, we create geometry of  breast in 3D and build a tetrahedral finite element mesh. Calculation is divided into two branches after initial and boundary conditions are declared. First branch is calculation of thermodynamic temperature based on equation (1). Second branch is calculation of electric field distribution in mammary gland according to equation (11). At the last stages, the brightness temperature is calculated using equation (10) and processing and analysis of obtained data is carried out.}
\end{minipage}
\end{figure}

To construct a stationary electric field distribution, it is convenient to solve the time-dependent Maxwell equations and as the result to obtain the stationary-state:
\begin{equation}\label{eq-Makswell}
    \frac{\partial \vec{B}}{\partial t} + rot(\vec{E}) = 0 \,, \quad \frac{\partial \vec{D}}{\partial t} - rot(\vec{H}) = 0 \,,\quad \vec{B}=\mu\vec{H}\,,\quad \vec{D}=\varepsilon\vec{E} \,,
\end{equation}
where  $\vec{B}$ is magnetic induction, $\vec{E}$ is electric field strength, $\vec{D}$ is electric induction, $\vec{H}$ is magnetic field strength, $\varepsilon(\vec{r})$ is the dielectric constant, $\mu(\vec{r})$ is magnetic permeability. Algorithm for approach to solving problem see fig.\ref{Algorithm}.

An important feature our models is variability of the physical parameters of biocomponents. This is due to a number of factors:
\begin{itemize}
\item a large individual variation in measured values due to gender, age and other differences;
\item the complexity object of measurement, ambiguity of the relationships between measured biophysical and corresponding biomedical characteristics;
\item the impossibility of measuring some characteristics of tissues without violating the integrity of organism;
\item small absolute values of measured values at high levels of internal noise.
\end{itemize}

Thus, we create a dataset that contains the results of computational experiments.

\section{Machine learning in analysis of microwave radiothermometry data}

\noindent This paper proposes a model for the existing method modernization for diagnosing breast cancer using microwave radiothermometry data. It is assumed that the method will be based on a hybrid technology founded on thermal fields mathematical modeling and thermometric data intelligent analysis. At the first stage, it is planned to measure temperatures in the IR and RTM ranges based on the existing standard methodology. Further, using Data Mining methods, the presumptive diagnosis and possible malignant neoplasms location are determined. At the next stage, using appropriate computer models, it is planned to measure internal and surface temperatures using a modified technique. New thermometric data will significantly improve the forecast for the temperature fields behavior inside the mammary glands. However, this approach places extremely high demands on the temperature data set quality obtained using computer simulation.

\begin{figure}[h]
\begin{center}
  \begin{minipage}{32pc}
\includegraphics[width=32pc]{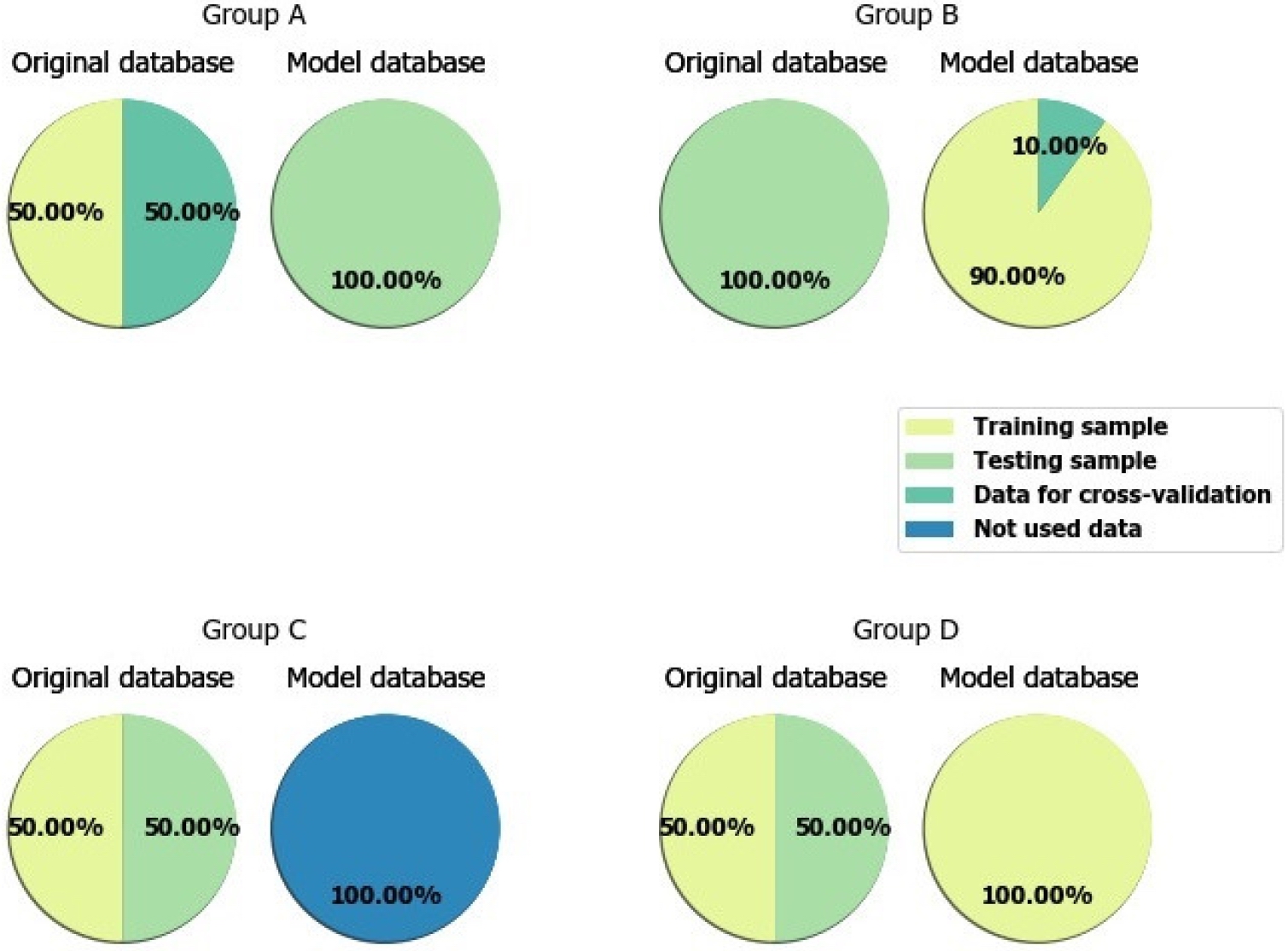}
\caption{\label{pie-charts}The structure of data set that are used for classification problem using machine learning methods.}
\end{minipage}
\end{center}
\end{figure}

Checking the model adequacy, in particular, its validation, was carried out as follows. Two thermometric databases were considered: the original database, i.e. obtained in cancer centers, and a model database obtained using computer modeling. The first includes thermometric data on 109 healthy right mammary glands (MG) and 27 patients with right breast cancer. The second includes thermometric data on 159 healthy and 160 patients with right breast. The choice of right MGs is solely due to the fact that a fairly large computer models set was calculated specifically for right MGs. Further, from these bases, four training and test samples groups were created. Group A training sample included half healthy and sick people from the original database, and the test sample included all the data from the model database. Group B training set included 90\% data from the model database, and the test set - all data from the original database. Group C training sample included half healthy and sick patients from the original database, all the rest were included in the test sample. Group D training set included half healthy and sick people from the original database and all model data, and the test set included the rest of the original database (see fig. \ref{pie-charts}). The classification was carried out using the following algorithms: K-nearest neighbors, naive Bayesian classifier, decision tree, random forest, logistic regression, gradient boosting, support vector machine. The following results were obtained. Sensitivity ($Sens$) reflects proportion of positive results that are correctly identified as such. Sensitivity of a diagnostic test indicates the likelihood that a sick subject will be classified as sick. Specificity ($Spec$) reflects proportion of negative results that are correctly identified as such. The geometric mean between sensitivity and specificity was used as a effectiveness ($eff$) measure. In addition, we present dispersion values ($dis$).

\begin{equation}\label{sens}
Sens=\frac{TP}{P}, Spec=\frac{TN}{N}, eff=\sqrt{Sens\cdot Spec},
\end{equation}
where $TP$ is number of sick patients identified by test, $P$ is number of real positive cases in data, $TN$ is number of healthy patients identified by test, $N$ is number of real negative cases in data.

\begin{table}[h!]
\caption{\label{blobs}Results of classification using various machine learning algorithms for created groups.}
\begin{center}
\begin{tabular}{lcccccccc}
\br
Classifier                & \multicolumn{2}{c}{Group A} & \multicolumn{2}{c}{Group B} & \multicolumn{2}{c}{Group C} & \multicolumn{2}{c}{Group D} \\
\mr
                          & $eff$          & $dis$          & $eff$          & $dis$          & $eff$          & $dis$          & $eff$          & $dis$          \\
K-nearest neighbors       & 0.56         & 0.06         & 0.6          & 0.03         & 0.45         & 0.07         & 0.63         & 0.08         \\
naive Bayesian classifier & 0.61         & 0.11         & 0.73         & 0.02         & 0.68         & 0.02         & 0.8          & 0.07         \\
decision tree             & 0.64         & 0            & 0.58         & 0.04         & 0.68         & 0.03         & 0.73         & 0.02         \\
random forest             & 0.62         & 0            & 0.59         & 0.04         & 0.63         & 0.01         & 0.74         & 0            \\
logistic regression       & 0.67         & 0.07         & 0.63         & 0.03         & 0.79         & 0.02         & 0.8          & 0.06         \\
gradient boosting         & 0.62         & 0.05         & 0.57         & 0.02         & 0.49         & 0.02         & 0.81         & 0.02         \\
support vector machine    & 0.33         & 0            & 0.31         & 0.01         & 0.79         & 0.13         & 0.8          & 0.02 \\
\br
\end{tabular}
\end{center}
\end{table}

In group A, the best results were obtained using logistic regression and decision tree. Their effectiveness measure is 0.67 and 0.64, respectively.

In group B, the best results were obtained using the naive Bayesian classifier (efficiency measure -- 0.73) and logistic regression (efficiency measure -- 0.63). In group C, the best results were obtained using logistic regression, where the effectiveness measure was 0.79. Sufficiently good results were shown by the support vector machine (efficiency measure -- 0.71), as well as a naive Bayesian classifier and decision tree (efficiency measure -- 0.68).

In group D, the best results were obtained using algorithms: gradient boosting (efficiency measure -- 0.81), support vector machine (efficiency measure -- 0.8), naive Bayesian classifier (efficiency measure -- 0.8) and logistic regression (efficiency measure -- 0.8). The general results are presented in Table \ref{blobs}.

It can be assumed that the efficiency measure in the first groups decreased in comparison with the works results \cite{Levshinskii3} and \cite{Losev2} due to a decrease in the feature space. Namely, the thermoasymmetry indicators, one of the most effective feature space elements, were not used. The latter poses the problem of developing paired organs computer models in aggregate. Signs based on pivot point and axillary temperatures were also not used. In addition, when compared with the results obtained on the neural networks basis, most of the history data (age, weight index, etc.) were not used.

\section{Conclusions and future work}
\noindent However, in general, the results obtained show high sufficiently computer models adequacy degree the obtained in \cite{Polyakov1, Levshinskii2, Polyakov2}, the possibility of their use in a hybrid method development for diagnosing breast cancer based on microwave radiothermometry data.

At present, most expert systems offer their solutions to doctor either in a deterministic form of an unambiguous conclusion or in form of probabilistic estimates of each possible diagnoses. A new approach to creation of diagnostic systems that explain decision made looks much more promising. More interest is development of advisory intelligent systems, that is expert systems containing a mechanism for explaining and justifying proposed solutions in a language understandable to user. Mathematical and computer modeling makes it possible to substantiate temperature anomalies in biological tissues. Development of a new method for diagnosis of breast diseases based on the combination original data and model data. Another direction of development is creation of paired models of mammary glands. Temperature asymmetry is an important diagnostic feature using microwave thermometry, therefore, modeling of both left and right breast will lead to the development of methods for diagnosing breast cancer.

\ack
MP  is grateful to RFBR according to the research project No. 19-37-90142 for the financial support in carrying out numerical modeling of the RTM-diagnostics process and assimilation of computer simulation data with original data, in particular, the method for modeling the thermal processes dynamics in mammary gland biological tissues has been modified, which provides the necessary solutions accuracy, stability, as well as a high convergence rate of the numerical method, caused by the needs of personalized medicine.
AL and AK are grateful to Russian Science Foundation (grant RFBR No. 19-01-00358) for the financial support of the development of mathematical models for early diagnosis of breast cancer, including the development of qualitative and quantitative characteristics complete system intended to justify the proposed diagnostic solution based on microwave radiothermometry data, on computational experiments and intelligent analysis of training thermometric data samples.

\end{document}